\documentclass[sigconf]{acmart}

\AtBeginDocument{%
  }

\setcopyright{acmlicensed}
\copyrightyear{2018}
\acmYear{2018}
\acmDOI{XXXXXXX.XXXXXXX}

\acmConference[Conference acronym xx]{ACM xx xx}{xx}{xx, xx}
\acmISBN{978-1-4503-XXXX-X/18/06}

\usepackage{microtype}
\usepackage{times} 
\usepackage{tabu}                      
\usepackage{lipsum}                    
\usepackage{mwe}                       
\usepackage{bbding}
\usepackage{pifont}
\usepackage{amsmath,amsfonts}
\usepackage{array}
\usepackage[caption=false,font=normalsize,labelfont=sf,textfont=sf]{subfig}
\usepackage{stfloats}
\usepackage{url}
\usepackage{verbatim}
\usepackage{graphicx}
\usepackage{bbm}
\usepackage{multirow}
\usepackage{comment}
\usepackage{color}
\usepackage{float}
\usepackage{makecell}

\setcopyright{none}
\renewcommand\footnotetextcopyrightpermission[1]{} 
\begin{document}

\title{Emotional Conversation: Empowering Talking Faces with Cohesive Expression, Gaze and Pose Generation}

\author{Jiadong Liang, Feng Lu}
\authornote{Corresponding author.}
\affiliation{%
  \city{State Key Laboratory of VR Technology and Systems, School of CSE, Beihang University}
  \state{Beijing}
  \country{China}}
\email{{ljdtc, lufeng}@buaa.edu.cn}


\begin{abstract}
Vivid talking face generation holds immense potential applications across diverse multimedia domains, such as film and game production. 
While existing methods accurately synchronize lip movements with input audio, they typically ignore crucial alignments between emotion and facial cues, which include expression, gaze, and head pose.
These alignments are indispensable for synthesizing realistic videos.
To address these issues, we propose a two-stage audio-driven talking face generation framework that employs 3D facial landmarks as intermediate variables. This framework achieves collaborative alignment of expression, gaze, and pose with emotions through self-supervised learning. 
Specifically, we decompose this task into two key steps, namely speech-to-landmarks synthesis and landmarks-to-face generation. 
The first step focuses on simultaneously synthesizing emotionally aligned facial cues, including normalized landmarks that represent expressions, gaze, and head pose. These cues are subsequently reassembled into relocated facial landmarks. In the second step, these relocated landmarks are mapped to latent key points using self-supervised learning and then input into a pretrained model to create high-quality face images. Extensive experiments on the MEAD dataset demonstrate that our model significantly advances the state-of-the-art performance in both visual quality and emotional alignment.
\end{abstract}

\ccsdesc[500]{Computing methodologies~Computer vision}

\keywords{Digital Human, Talking Face Generation, AIGC}
\begin{teaserfigure}
    \centering
    \includegraphics[width=1.0\textwidth]{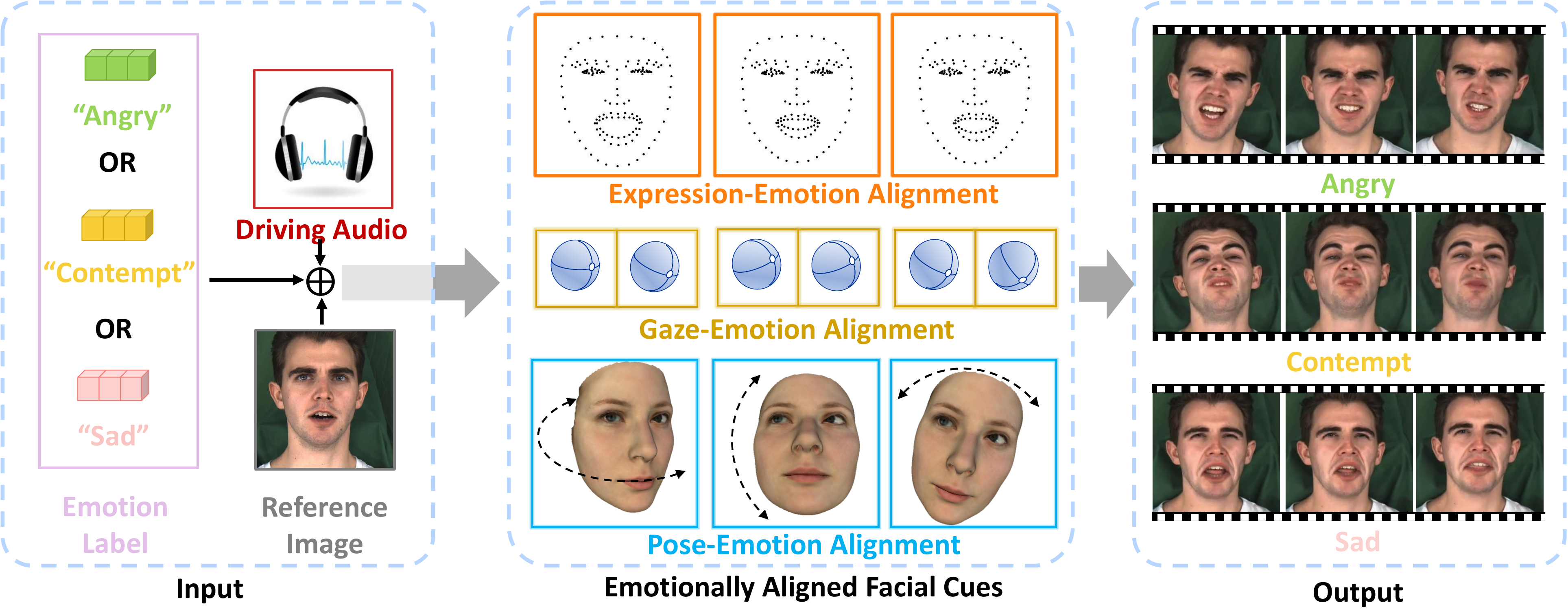}
    \caption{We propose an advanced two-step framework for synthesizing vivid emotional talking faces with emotionally aligned facial cues. Initially, our approach utilizes the provided driving audio and an emotion label to generate three sequences of fine-grained facial cues (expression, head pose, and gaze) tailored to the specified emotion. Subsequently, these facial cues align with the specified emotion through self-supervised learning. Finally, utilizing these facial cues, we can produce vivid emotional talking face videos.}
    \label{fig:teaser}
\end{teaserfigure}

\received{20 February 2007}
\received[revised]{12 March 2009}
\received[accepted]{5 June 2009}

\maketitle
\section{Introduction}
\begin{table*}[t]
\begin{center}
  \caption{A comparison of related works across various criteria is shown on the left.}
  \vspace{-6pt}
  \label{tab:prior_work}
  \scriptsize%
	\centering%
\resizebox{1.0\linewidth}{!}{
  \begin{tabular}{l|cccccccccccc}
  \toprule
   Facial Cues & \textbf{Ours} &  Wave2Lip  & MakeItTalk & PC-AVS & Audio2Head & MEAD & EVP & EAMM & EMMN & Xu & PD-FGC & SPACE \\
  \midrule
   Landmarks  & \ding{108} & \ding{109} & \ding{108} & \ding{109} & \ding{109} & \ding{109} & \ding{109} & \ding{109} & \ding{109} & \ding{109} & \ding{109} & \ding{108}\\
   Head Pose  & \ding{108} & \ding{119} & \ding{108} & \ding{119} & \ding{119} & \ding{109} & \ding{119} & \ding{109} & \ding{109} & \ding{109} & \ding{119} & \ding{108}\\
   Gaze       & \ding{108} & \ding{109} & \ding{109} & \ding{109} & \ding{109} & \ding{109} & \ding{109} & \ding{109} & \ding{109} & \ding{109} & \ding{119} & \ding{119}\\
   Emotion    & \ding{108} & \ding{109} & \ding{109} & \ding{109} & \ding{109} & \ding{108} & \ding{108} & \ding{108} & \ding{108} & \ding{108} & \ding{119} & \ding{108}\\
  \bottomrule
  \end{tabular}}
  \end{center}
\vspace{-3mm}
\end{table*}

The task of talking face generation involves creating a video of a talking face using a still identity image of the speaker and an audio track of their speech content. 
Furthermore, the generation of emotional talking face videos, featuring precise lip synchronization and vivid facial cues such as expressions, gaze, and head pose holds considerable potential for future applications.
We found that these facial cue sequences exhibit consistent patterns across videos corresponding to specific emotions. For instance, in a talking face depicting contempt, individuals typically narrow their eyes, tilt their heads upward, and shift their gaze horizontally. Conversely, in videos portraying surprise, individuals generally widen their eyes while maintaining a forward-facing head pose and gaze.
The alignment of these facial cues with emotions is crucial for synthesizing realistic talking face videos.
Therefore, the primary challenge of this task is to produce realistic facial videos that not only accurately reproduce lip movements in response to the driving audio but also maintain consistency between the various facial cues and corresponding emotions.

With the development of AI-Generated Content (AIGC), numerous advanced methods for generating emotional talking face have emerged. Corresponding methods can be mainly divided into audio-driven talking face generation \cite{MEAD, EAMM, space} and video-driven talking face generation \cite{EVP, PDFC}.
Audio-driven talking face generation synthesizes a sequence of portrait images by using both an identity image and the corresponding audio as inputs.
However, most existing works focus exclusively on lip movement and often lack the ability to generate associated facial cues.
The latest work, SPACE \cite{space}, is a multi-stage generative framework that achieves fine-grained control over facial expressions, emotion categories, head poses, and gaze directions of generated faces by manipulating the intermediate landmarks.
Unfortunately, in SPACE, the control and generation of gaze and head pose sequences are unrelated to emotion. Although the generated faces display high visual quality, the final videos lack vividness and fail to differentiate emotions effectively.
Video-driven talking face generation involves using a single identity image and multiple driving source videos as inputs. By employing a contrastive learning strategy, relevant features are extracted from various source videos to generate high-quality target talking face videos. These methods allow for the editing of facial attributes, such as head pose, facial expression, gaze, blinking, and audio, by modifying the corresponding source videos. 
However, these methods are not only expensive in the training stage but also incur additional costs in the inference stage due to the need for source video collection.
We find that the fine-grained facial cues of the talking face video, such as expression, gaze and head pose, are closely related to emotional categories. Independently controlling these facial cues without considering emotion can lead to unrealistic results. Therefore, we require an audio-driven method that combines the low cost of audio-driven methods with the high alignment between facial cues and emotion to synthesize vivid emotional talking face videos.

In this paper, we propose an advanced audio-driven method to synthesize vivid emotional talking faces with emotionally aligned facial cues. We decompose the task into two sub-tasks: speech-to-landmarks synthesis and landmarks-to-face generation. 
Given input speech, an emotion label, and an identity image, the proposed speech-to-landmarks module can generate sequences of normalized facial landmarks (representing expressions), gaze, and head pose in an auto-regressive manner. 
To address the issue of substantial variations in gaze, we discretized the eye region and modeled gaze prediction as a classification task, successfully predicting gaze sequences for the first time.
The alignment of emotional labels with these facial cues is achieved through self-supervised learning.
These generated facial cues are synthesized into relocated 3D facial landmarks through coordinate correction and rotation. The relocated facial landmarks, driven by these cues, not only synchronize with the input audio but also enhance the alignment of expression, gaze, and pose movements with the corresponding emotion labels.
We also build a collaborative emotion classifier to model the intrinsic relationships among these facial cues.
Specifically, the classifier takes aggregated intermediate features from facial cues as input ensuring consistency among these facial cue sequences.
The proposed landmarks-to-face module utilizes a latent keypoints space\cite{siarohin2021motion}, capable of producing more realistic faces compared to traditional facial landmarks. Specifically, this module maps the relocated landmarks, generated by the speech-to-landmarks, to latent feature points. It then employs a pre-trained generator\cite{facevid2vid} to synthesize high-quality facial images.

\textbf{To the best of our knowledge, this is the first attempt to align normalized facial landmarks, gaze, and head pose concurrently with emotional categories. Compared to existing works, these explicitly aligned facial cues  significantly improve the intensity and accuracy of emotional expressions in generated talking faces.}
Additionally, other contributions are as follows:
\begin{itemize}
\item We introduced an advanced speech-to-landmarks synthesis auto-regressive algorithm. This algorithm can generate emotionally aligned facial cues, including normalized facial landmarks, gaze, and head pose, in an auto-regressive manner.
\item We designed a specific eye region discretization strategy that efficiently generates gaze sequences by modeling gaze prediction as a classification task.
\item Extensive experiments on the MEAD\cite{MEAD} dataset demonstrate the superiority of the proposed method in terms of lip synchronization and the quality of generated faces.
\end{itemize}

\section{Related Work}
In the following section, we provide an overview of prior research on audio-driven talking face generation and video-driven talking face generation.
Table \ref{tab:prior_work} outlines a summary of  the key differences between our approach and other state-of-the-art methods. 
Specifically, a solid circle indicates that the method can automatically generate the relevant attribute or be driven by other source videos to control the relevant attribute.
A half-filled circle denotes that the related attribute can only be controlled by other source videos. Conversely, an empty circle represents  that the method cannot generate or control the related attribute.

\noindent\textbf{Audio-driven Talking Face Generation.}
The objective of Speech-driven Talking Face Generation \cite{karras2017audio, taylor2017deep, wang2021audio2head,zhou2019talking,vougioukas2018end,liu2022semantic,wang2023lipformer,shen2023difftalk,fan2022faceformer,du2023dae,wu2023speech} is to establish a mapping from the input speech to facial representations.
MakeItTalk\cite{zhou2020makelttalk} disentangles audio content and speaker information to control lip motion and facial expressions, and effectively works with various portrait styles.
Audio2Head\cite{wang2021audio2head} addresses challenges in achieving natural head motion by employing a motion-aware RNN for head pose prediction.
With the further development of this field, there is a growing emphasis on controlling facial emotions in generated talking faces.
Wang collected the MEAD\cite{MEAD} dataset and proposed a method that conditions talking head generation based on emotion labels. However, MEAD primarily focuses on controlling only the mouth region while leaving other parts unchanged, leading to a lack of continuity in the generated videos.
Compared to MEAD, which uses a single emotion label as input to control the generation of video emotion categories, EAMM\cite{EAMM} achieves precise emotional control over the synthesized video by adopting features extracted from the emotion source video. However, the method still ignores the movement of the gaze direction and head pose, which results in less realistic generated videos.
EMMN\cite{tan2023emmn} and \cite{xu2023high} employ memory networks and textual prompts, respectively, to control the emotions in the generated videos.
SPACE\cite{space} achieves fine-grained control over facial expressions, emotion categories, head poses, and gaze directions of generated faces by decomposing the generation task into multiple subtasks.
While these methods are user-friendly and straightforward, they often struggle to generate emotionally aligned facial cues for generated videos.

\noindent\textbf{Video-driven Talking Face Generation.}
The goal of video-driven talking face generation\cite{drobyshev2022megaportraits, gu2020flnet, hong2022depth, wang2022latent,burkov2020neural, liu2021li, xiang2020one} is to accurately map facial movements from a source video onto a target image.
Wave2Lip\cite{prajwal2020lip} constructs an expert discriminator to ensure precise alignment between the lip movements and the input audio.
PC-AVS\cite{zhou2021pose} utilizes an implicit low-dimension pose code to separate audio-visual representations to achieve accurate lip-syncing and pose control.
EVP\cite{EVP} decomposes speech into two decoupled spaces to generate dynamic 2D emotional facial landmarks.
PD-FGC\cite{PDFC} allows for the editing of facial attributes, such as head pose, facial expression, gaze, blinking, by modifying the
corresponding source videos. 
These methods are not only costly during the training stage, but also entail additional expenses during the inference stage due to the requirement of source video collection.
\begin{figure*}[t]
    \centering
    \includegraphics[width=\textwidth]{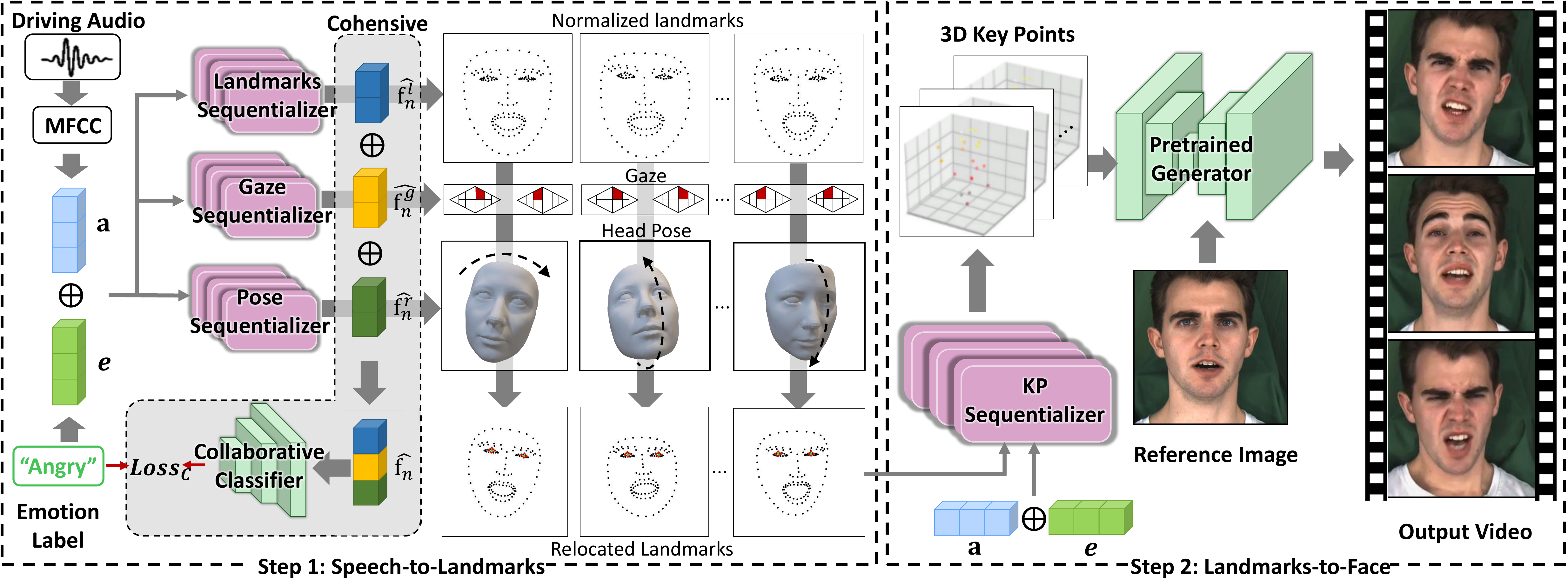}
    \caption{Architecture of the proposed method, which performs emotional talking face generation in two steps. In Step 1, we innovatively achieved the simultaneous generation of facial cue sequences, including normalized landmarks, gaze, and head pose. These cues are then aligned with emotional labels via self-supervised learning. In Step 2, we utilize the emotionally aligned facial cues from Step 1 as inputs, employing a pre-trained model to produce vivid emotional talking face videos.}
    \vspace{-3mm}
    \label{fig:framework}
\end{figure*}

\section{Two-stage Talking Face Generation}
\begin{figure}[t]
    \centering
    \includegraphics[width=\columnwidth]{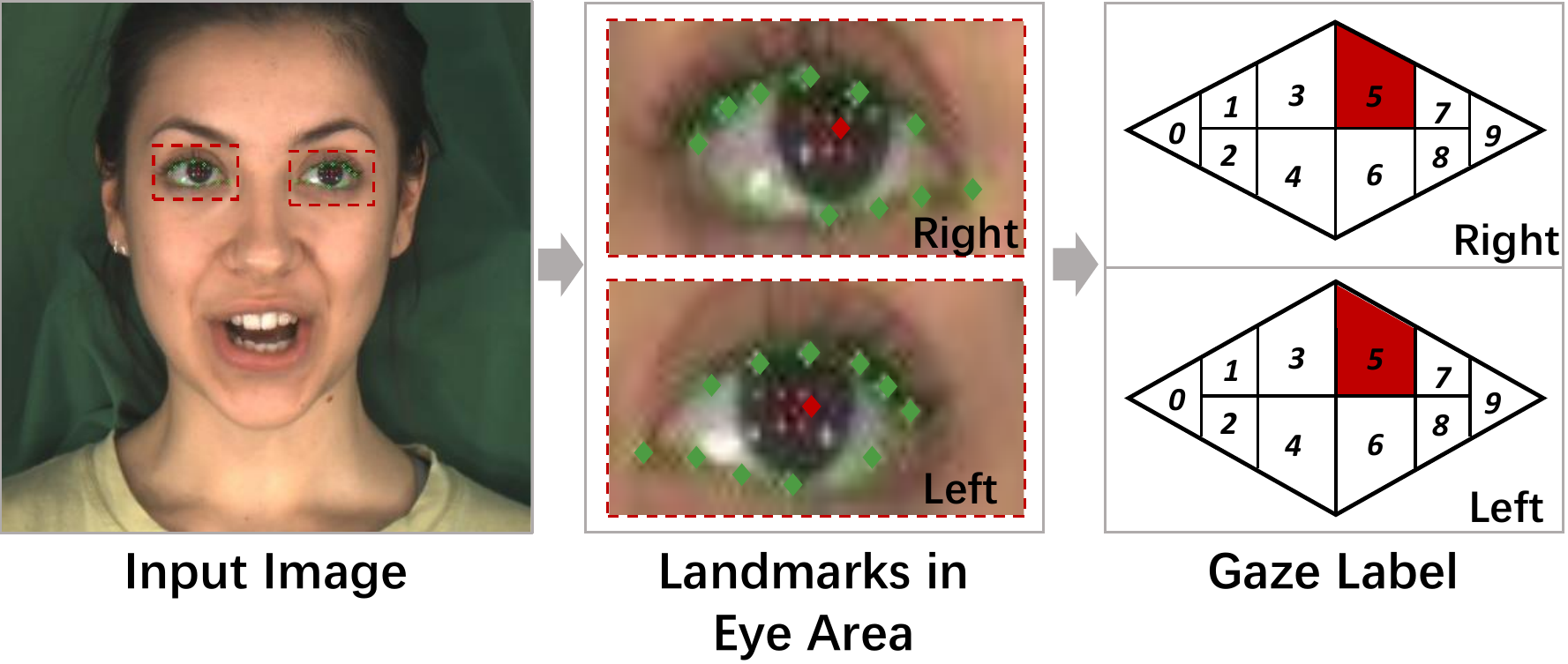}
    \caption{Pipeline of gaze direction discretization.}
    \vspace{-5mm}
    \label{fig:gaze_dis}
\end{figure}

The proposed method takes the driving audio and a reference image, along with an emotion label, and produces an emotional talking face. It decomposes this task into two steps:
(1) Speech-to-Landmarks Synthesis: Given a reference image, it extracts normalized landmarks, gaze labels, and head poses, and predicts their per-frame motions driven by the input speech and emotion label.
Specifically, our proposed method innovatively generates cohesive sequences of normalized facial landmarks, gaze, and head pose simultaneously.
Furthermore, we accomplish the collaborative alignment of these facial cues with the corresponding emotional labels using self-supervised learning.
(2) Landmarks-to-Face Generation: In this step, the per-frame relocated facial landmarks are mapped to latent keypoints, which are then fed into the pre-trained model \cite{facevid2vid} to generate the final emotional talking face.
This decomposition offers multiple advantages. Firstly, it enables fine-grained control over the output facial expressions. Secondly, the two-stage training approach can reduce the training complexity and accelerate the convergence of each module. By leveraging a pretrained face generator, we can effectively reduce the training cost while obtaining high-quality emotional talking faces. The overall framework is shown in Fig. \ref{fig:framework}.

\subsection{Speech-to-Landmarks Synthesis.}
Given an input speech, emotion label, and an identity face image, the proposed speech-to-landmarks synthesis is capable of generating sequences of normalized facial landmarks (representing expressions), gaze, and head pose in an auto-regressive manner. 
These facial cues are further aligned with specific emotions through self-supervised learning.
Specifically, the network for speech-to-landmarks synthesis consists of three modules: Landmark Sequentializer, Gaze Sequentializer, and Pose Sequentializer, each performing auto-regressive prediction on different facial cues sequences.

\noindent\textbf{Landmark Sequentializer.}
In talking face generation, the quality of facial landmark generation is paramount as these landmarks directly control lip movements and expressions.
Given the input audio MFCC features and the normalized 3D facial landmarks of the input image. This is represented by
\begin{equation}
\begin{split}
\mathbf{f}_n^l = \mathcal{S}_{\text{landmark}}(\mathbf{C}_{n-1}, \mathbf{a}_n, \mathbf{e}), \quad \mathbf{C}_n = \text{Linear} (\mathbf{f}_n^l), 
\label{eqn:lstm_lms)}
\end{split}
\end{equation}
where $\mathbf{f}_n^l$ and $\mathbf{a}_n$ are the facial landmarks features and audio features at time $n$, respectively.
$\mathcal{S}_{\text{landmark}}$ represents the auto-regressive generator, and $\mathbf{C}_n \in \mathbb{R}^{147 \times 3}$ is the normalized 3D facial landmarks at time $n$.
The input emotion label embedding $\mathbf{e} \in \mathbb{R}^D$ corresponds to the indexed vectorial representation in the embedding matrix $\mathbf{E} \in \mathbb{R}^{D \times K}$ for an emotion dictionary containing $K$ emotion categories, which is learned together with the whole model.
We employ a convolutional neural network (CNN) to encode the audio data, while a multi-layer perceptron (MLP) is used to encode the 3D facial landmarks. The network is trained utilizing an L1 loss function, which reduces the L1 distance between the predicted facial landmarks and the normalized ground truth landmarks. Notably, we assign a higher loss scale to the y-axis to emphasize vertical motion errors during training \cite{space}. 
Using facial landmarks as an intermediary representation is advantageous as it facilitates the explicit manipulation of facial features. For instance, by manipulating the landmarks of the eyes, it becomes possible to incorporate eye blinks into the face. We discovered that conducting predictions in the normalized space is crucial to simplify the mapping between phonemes and lip motions.

\noindent\textbf{Pose Sequentializer.}
The natural movement of the head can effectively enhance the vividness of the generated video, but its motion pattern is also influenced by the emotional category. 
Hence, we also train an auto-regressive $\mathcal{S}_{headpose}$ that predicts the rotation and translation for the facial landmarks. The rotation is represented by three angles: yaw, pitch, and roll, and the corresponding translation is the displacement of the 3D landmarks in the x-axis, y-axis, and z-axis.
The prediction is represented by
\begin{equation}
\resizebox{0.9\linewidth}{!}{$
\begin{aligned}
\mathbf{r}_n = \text{Liner}(\mathbf{f}_n^r), \quad \mathbf{r}_n = [\textbf{m}_n, \textbf{b}_n], \quad \mathbf{f}_n^r = \mathcal{S}_{headpose}(\mathbf{r}_{n-1}, \mathbf{a}_n, \mathbf{e}),
\label{eqn:lstm_pose)}
\end{aligned}
    $}
\end{equation}
where $\textbf{m}_n = [yaw, pitch, roll]$, $\textbf{b}_n = [\Delta_x, \Delta_y, \Delta_z]$
and $\mathcal{S}_{\text{headpose}}$ represents the pose sequentializer.
The poses, whether predicted or extracted from a reference video, are applied to the frontal normalized landmarks predicted by our Landmark Sequentializer. This transformation maps the normalized landmarks back to the image space after applying an appropriate scaling factor.
The Pose Sequentializer is also trained utilizing an L1 loss.

\noindent\textbf{Gaze Sequentializer.}
The eyes, as important organs for human interaction, contain abundant information in gaze direction, which can impact the emotion category of generated videos. In contrast to the prediction of facial landmarks and head pose, we transform the prediction of gaze direction from regression to classification by discretizing the eye regions as shown in Fig \ref{fig:gaze_dis}. It effectively enhances the accuracy of gaze direction prediction.
However, in the process of predicting the sequence of gaze directions, the current gaze direction heavily relies on the previous gaze. 
\begin{figure}[t]
    \centering
    \includegraphics[width=1.0\columnwidth]{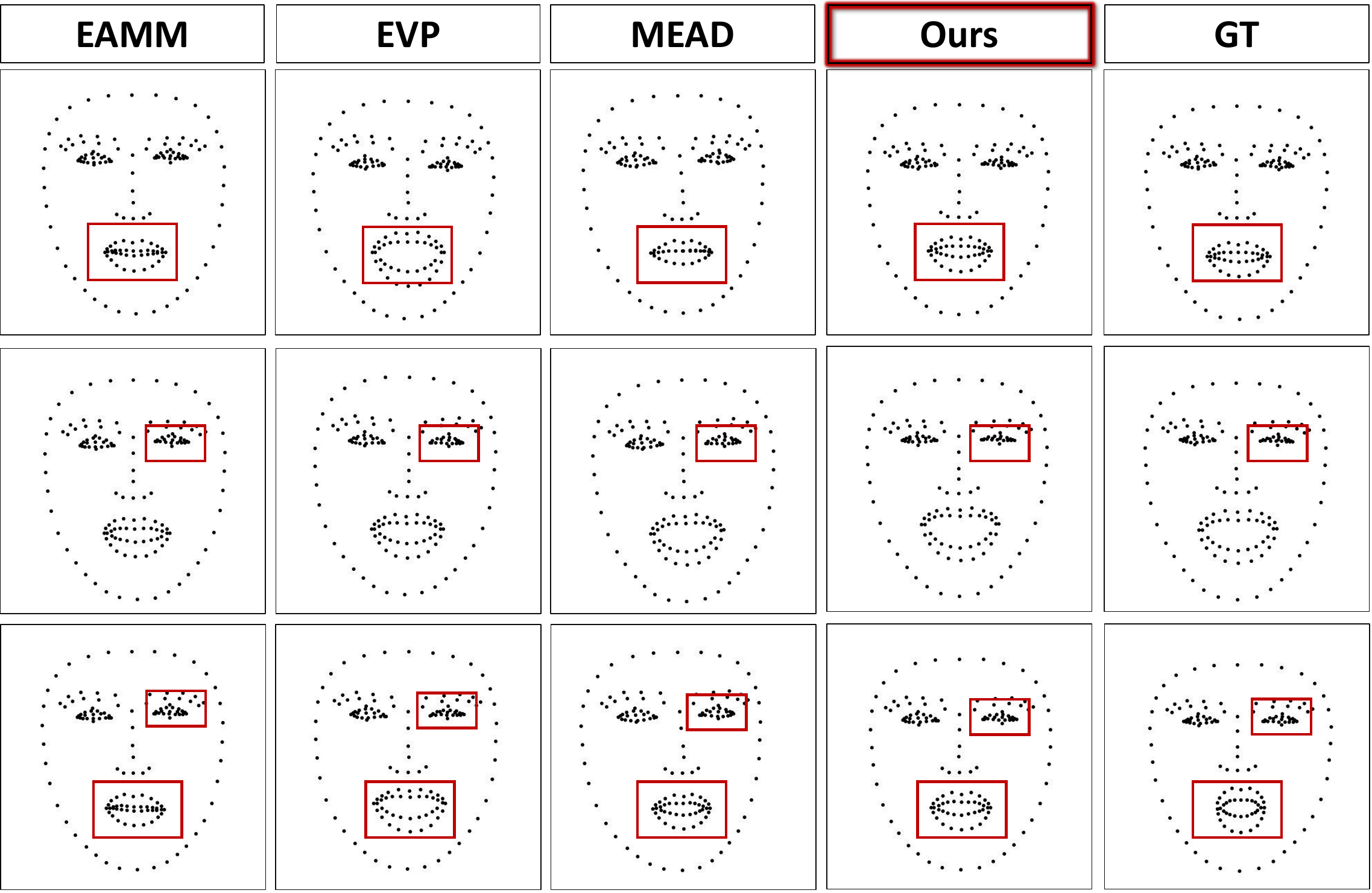}
    \caption{Qualitative comparison of generated normalized landmarks between our method and three other methods on the MEAD dataset.}
    \vspace{-5mm}
    \label{fig:lms_compare}
\end{figure}
We adopt $\mathcal{S}_{gaze}$ to model this dependency relationship. 
Specifically, the gaze prediction of two eyes $[\mathrm{u}_n^{left} \in [0, S-1], \mathrm{u}_n^{right} \in [1, S-1]]$ at time n can be modeled as:
\vspace{-2mm}
\begin{equation}
\begin{split}
\mathbf{f}_n^g = \mathcal{S}_{gaze}(\mathrm{v}_{n-1}, \mathbf{a}_n, \mathbf{e}). 
\label{eqn:lstm_gaze)}, \quad \mathrm{v}_n = \mathrm{u}_n^{left} + S \times \mathrm{u}_n^{right}, 
\end{split}
\end{equation}
Formally, gaze decoder performs classification at $n$-th time step based on the hidden states $\mathbf{f}_n^g$ by
\begin{equation}
\begin{split}
    v_n = \underset{i \in [1, S \times S]}{\text{argmax}(\mathbf{p}_n^i)}, \quad \mathbf{p}_n = \text{Softmax}(\mathbf{M} \mathbf{f}_n^g), 
    \label{eqn:cls}
\end{split}
\end{equation}
where $\mathbf{M}$ denotes a linear transformation. $\mathbf{p}_n \in \mathbb{R}^{S \times S }$ represents calculated probabilities for a total of $S \times S$ classification entries. $v_n$ is the entry with the maximal probability, from which we can infer the corresponding gaze label $[\mathrm{u}_n^{left}, \mathrm{u}_n^{right}]$.
Finally, we utilize the cross-entropy loss to optimize the gaze direction prediction model.
\begin{equation}
\begin{split}
    \mathrm{Loss}_{gaze} =  \frac{1}{N}\sum_{n=1}^N \mathrm{Loss}_{\text{CE}}(\mathbf{p}_n, \hat{\mathbf{p}_n}).
\label{eqn:cross}
\end{split}
\end{equation}
After obtaining these three types of facial cues, we integrate the gaze and head pose data into the normalized landmarks to obtain the relocated landmarks as shown in Fig. \ref{fig:framework}.
We also investigate the inherent relationships within these facial cues in the specific emotion.
Therefore, we construct a collaborative emotion classifier to push consistency among these facial cue sequences. 
The classifier predicts the emotion category using the aggregated intermediate features from facial cues as input.
Specifically, we can obtain the fake intermediate features $\hat{\mathbf{f}_n} = [\hat{\mathbf{f}_n^l}; \hat{\mathbf{f}_n^r}; \hat{\mathbf{f}_n^g}]$, where $[\hat{\mathbf{f}_n^l}; \hat{\mathbf{f}_n^r}; \hat{\mathbf{f}_n^g}]$ is predicted during the training stage.
\begin{align}
\begin{split}
\hat{\mathbf{l}_n} = \mathcal{F}_{classify}(\hat{\mathbf{f}_n}), \quad \mathrm{Loss}_{C} =  \mathrm{L}_{\text{CE}}(\hat{\mathbf{l}_n}, \mathbf{l}_n)
\end{split}
\end{align}
where $\mathrm{L}_{\text{CE}}$ is cross entropy loss and $\mathbf{p}_t$ is calculated probabilities for total emotion classification entries.
The total loss for the stage of speech-to-landmarks synthesis is 
\begin{equation}
\resizebox{0.9\linewidth}{!}{$
\begin{aligned}
\mathrm{Loss}_{norm} = 
\quad \mathrm{Loss}_{landmarks} + \mathrm{Loss}_{pose} + \mathrm{Loss}_{gaze} + \mathrm{Loss}_{C}. 
\label{eqn:lms_all_loss}
\end{aligned}$}
\end{equation}
\begin{figure*}[t]
    \centering
    \includegraphics[width=1.0\textwidth]{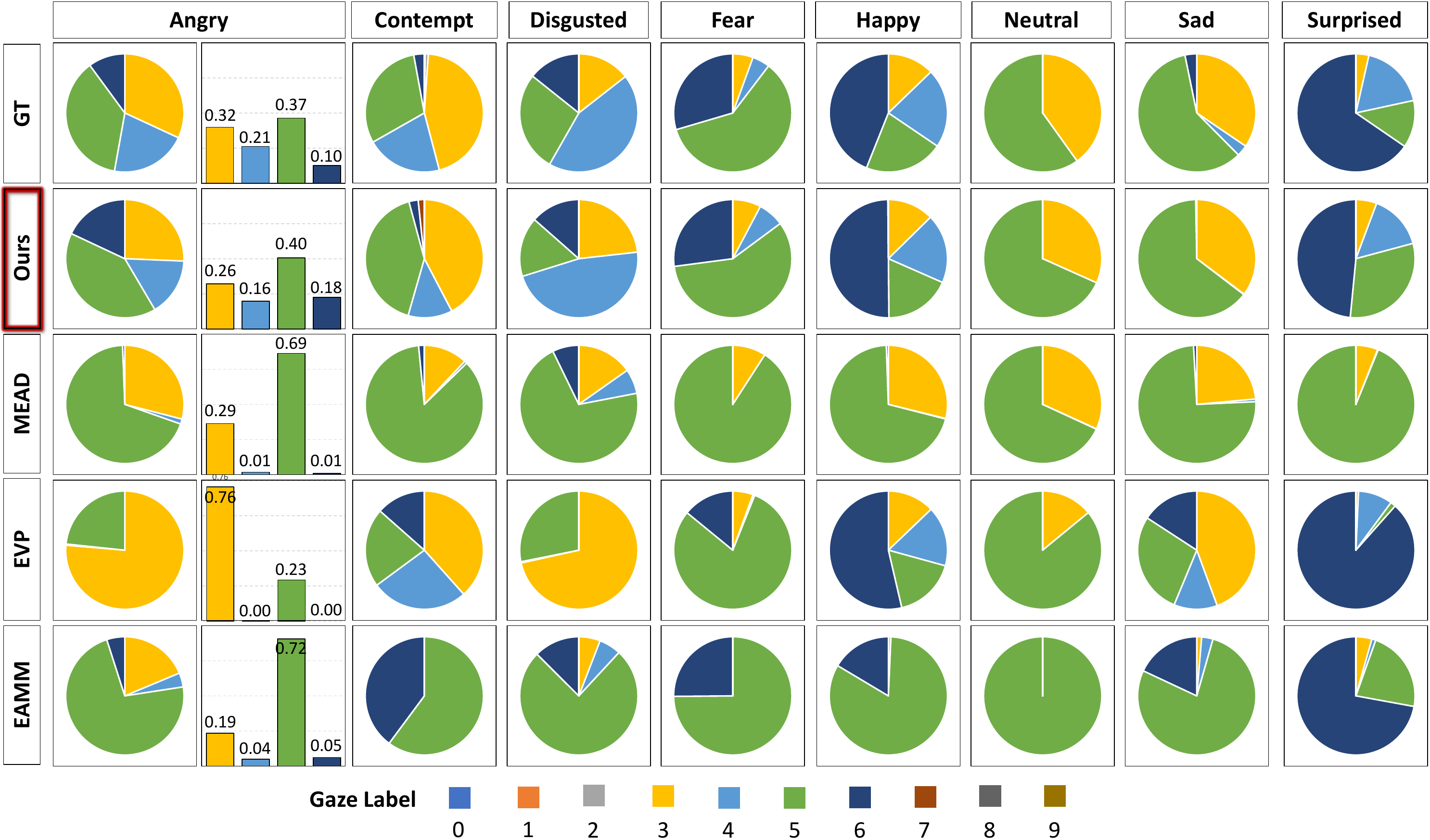}
    \caption{Comparison of left eye gaze distribution between our model and state-of-the-art models across different emotion categories.}
    \vspace{-2mm}
    \label{fig:gaze_res}
\end{figure*}
In fact, the generation of all facial cues incorporates emotional labels as input. By minimizing the discrepancy between the generated facial cues and the ground truth detected by off-the-shelf detectors (see Sec.\ref{sec:data_preprocess}), we achieve alignment of expressions, gazes, and head poses with emotion labels through self-supervised learning.

\subsection{Landmarks-to-Face Generation.}
The field of face generation has undergone rapid development, we find that using latent keypoints as input can generate high-quality facial images. 
Following SPACE\cite{space}, we utilized the pre-trained model face-vid2vid \cite{facevid2vid}, a state-of-the-art framework for generating faces from latent keypoints. 
This approach avoids the need to learn a facial image generator from scratch, thereby reducing computational requirements and improving efficiency.
Specifically, we first use the relocated landmarks in the 3D space generated in the speech-to-landmarks synthesis stage, along with the input speech and emotion labels, as input to perform autoregressive prediction of latent keypoints in a self-supervised learning manner, as shown in Fig. \ref{fig:framework}. 
\begin{figure*}[t]
    \centering
    \includegraphics[width=1.0\textwidth]{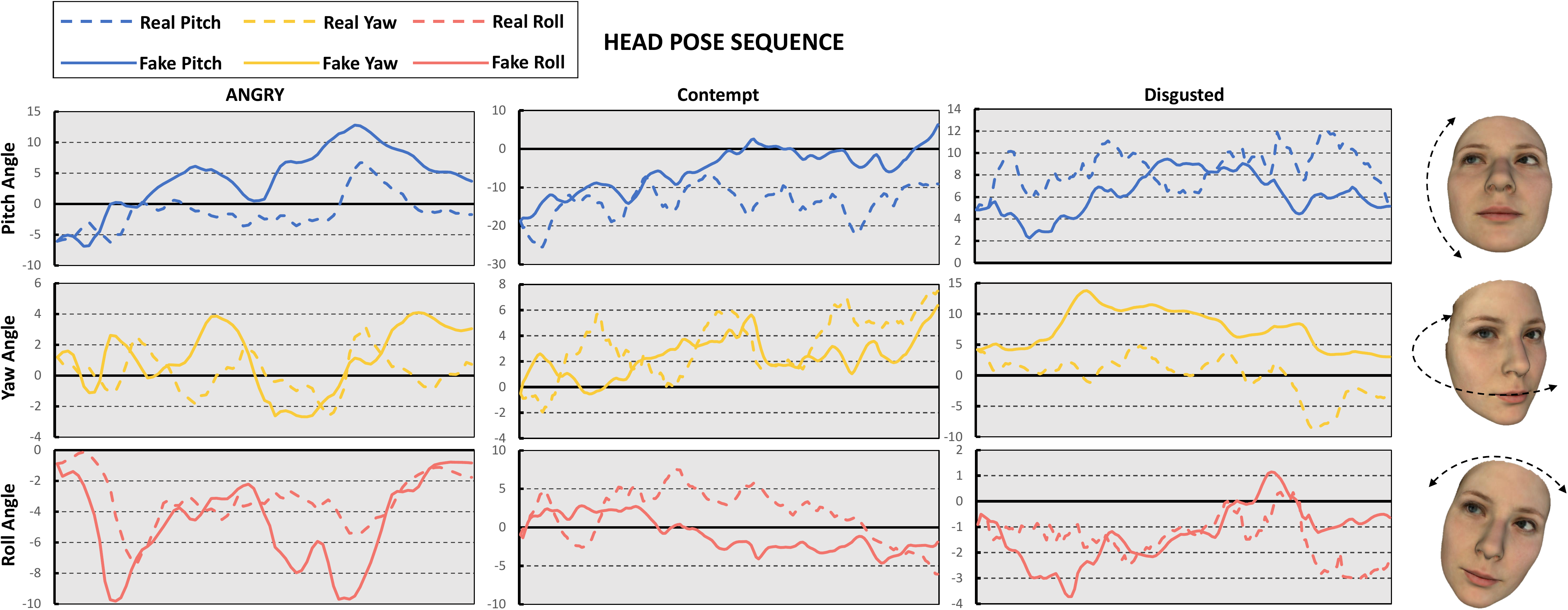}
    \caption{Visualization of the head pose sequences in the pitch, yaw, and roll directions under different emotions.}
    \vspace{-3mm}
    \label{fig:he_res}
\end{figure*}
\begin{equation}
\resizebox{.65\hsize}{!}{$
\begin{aligned}
& \mathbf{R}_n = \mathcal{F}_{relocate}(\mathbf{C}_{n} \cdot \mathbf{m}_{n} + \mathbf{b}_{n}, \mathrm{u}_n^{left}, \mathrm{u}_n^{right}), \\
& \mathbf{K}_n = \mathcal{S}_{Key}(\mathbf{K}_{n-1}, \mathbf{R}_{n}, \mathbf{a}_n, \mathbf{e}), \\
\label{eqn:lstm_key)}
\end{aligned}$}
\vspace{-3mm}
\end{equation}
where the $\mathcal{F}_{relocate}$ converts the normalized 3D facial landmarks $\mathbf{C}_n \in \mathbb{R}^{147 \times 3}$  to relocated facial landmarks $\mathbf{R}_n \in \mathbb{R}^{147 \times 3}$  using the gaze label $[\mathrm{u}_n^{left}, \mathrm{u}_n^{right}]$ and head pose $\mathbf{r}_n = [\textbf{m}_n, \textbf{b}_n]$ obtained in the stage of speech-to-landmarks.
$\mathcal{S}_{Key}$ is auto-regression generator for 3D key points.
Then, given the predicted latent keypoints $\mathbf{K}_n \in \mathbb{R}^{10 \times 3}$ and the initial face, the pre-trained model generates high-quality facial images by using a flow-based warping field as intermediate variables.
Finally, we can generate high-quality facial images with a resolution of 256, which is generally superior to previous works. 
By breaking down the generation of 3D facial landmarks into the collaborative production of three facial cues, we have simplified the task. Consequently, we selected a lightweight Bi-LSTM as the backbone for all auto-regressive generators ($\mathcal{S}_{lanmark}$, $\mathcal{S}_{headpose}$, $\mathcal{S}_{gaze}$, and $\mathcal{S}_{Key}$), enabling the generation of high-quality facial cues and 3D keypoints with minimal computational expense.

This work fundamentally differs from SPACE \cite{space}. SPACE focuses on precise control of the generated faces but neglects the alignment of related facial cues with emotions. 
Consequently, this oversight results in low differentiation among emotion categories and diminished vividness in the generated emotional talking faces.
Conversely, our research prioritizes maintaining consistency between the facial cues and the driving emotion labels in generated emotional talking face videos. Furthermore, utilizing our proposed eye region discretization strategy, we have successfully generated gaze sequences for the first time.

\section{Experiment}
\subsection{Implementation Details.}
The videos were sampled at a rate of 30 frames per second (FPS), while the audio was pre-processed to 16 kHz. To extract audio features, we computed 28-dimensional MFCC using a window size of 30.
We trained and evaluated our method using the MEAD dataset, an audio-visual emotional dataset comprising 60 actors/actresses and eight different emotion categories.

\subsection{Dataset Preprocessing.}
\label{sec:data_preprocess}
The proposed method adopts a self-supervised learning approach in both Speech-to-Landmarks Synthesis and Landmarks-to-Face Generation. To achieve this, we preprocess the emotional talking face videos to obtain training pairs by extracting the facial landmarks and latent keypoints for each frame.
Variations in head poses within videos lead to a greater diversity of landmark movements, potentially diminishing prediction accuracy. Consequently, our work performs face normalization. Specifically, videos are aligned by centering on the nose tip and resized to a uniform resolution of 256 × 256 pixels, a data processing approach prevalent in related works. \cite{chen2020talking, chen2018lip, zhou2019talking}.
Given a talking-head video, we first extract per-frame facial landmarks and head poses. 
We adopt the Mediapipe \cite{lugaresi2019mediapipe} landmark detector to extract 478 3D facial landmarks from each frame.To reduce computational costs and enhance model inference speed, we have selected a total of 147 facial landmarks for training while retaining the landmarks in the eye area.
The per-frame head pose is obtained by the 3DDFA \cite{guo2020towards} landmark detector.
We then rotate the face such that the nose tip faces straight towards the camera, aligned with the camera axis.
The per-frame frontalized 3D facial landmarks will be scaled to obtain normalized landmarks with the same facial width.

The majority of existing speech-driven talking face generation methods have neglected the generation of gaze direction sequences, resulting in individuals in the generated videos maintaining their gaze forward. This omission weakens the authenticity of the generated videos.
In this paper, to achieve more effective prediction of gaze direction, we discretize gaze direction and model the prediction as a classification problem rather than a regression problem. 
Specifically, we divide each eye area into 10 regions based on facial landmarks. During the training process, we directly predict into which region the pupil will fall. The pipeline of gaze discretization is illustrated in Fig. \ref{fig:gaze_dis}."
In addition to landmarks, we also extract latent keypoints per frame using the pretrained face-vid2vid encoder \cite{facevid2vid}. 
This provides per-frame pairs of (facial landmarks, latent keypoints).

\subsection{Evaluation Metrics.}
To evaluate the alignment between the generated face and the input audio, we  calculate the Euclidean distance of facial landmarks between the generated images and the ground truth images in the mouth region (MLD \cite{chen2019hierarchical}). We also evaluate the accuracy of facial expressions by measuring the landmarks difference on the whole face (FLD).
We use the confidence scores of SyncNet \cite{chung2017out} to evaluate the consistency between the generated face and the driving audio at the feature level.
For the visual quality of the synthesized face, we use Structural Similarity (SSIM), Peak Signal to Noise Ratio (PSNR), and  Frechet Inception Distances (FID) \cite{heusel2017gans} for quantitative analysis of the generated results.
Additionally, we need to conduct a quantitative assessment of the effectiveness of gaze and head pose in talking face videos. Given the diversity of these facial cues, calculating their frame-by-frame differences from the ground truth is not meaningful. Therefore, we adopt Dynamic Time Warping (DTW) to measure the similarity between the generated facial cues sequences and GT. Specifically, for head pose, we separately calculate the DTW for Pitch, Yaw, and Roll sequences. For gaze, we calculate the DTW for the pupil movement speed sequences.

\begin{figure*}[t]
    \centering
    \includegraphics[width=1.0\textwidth]{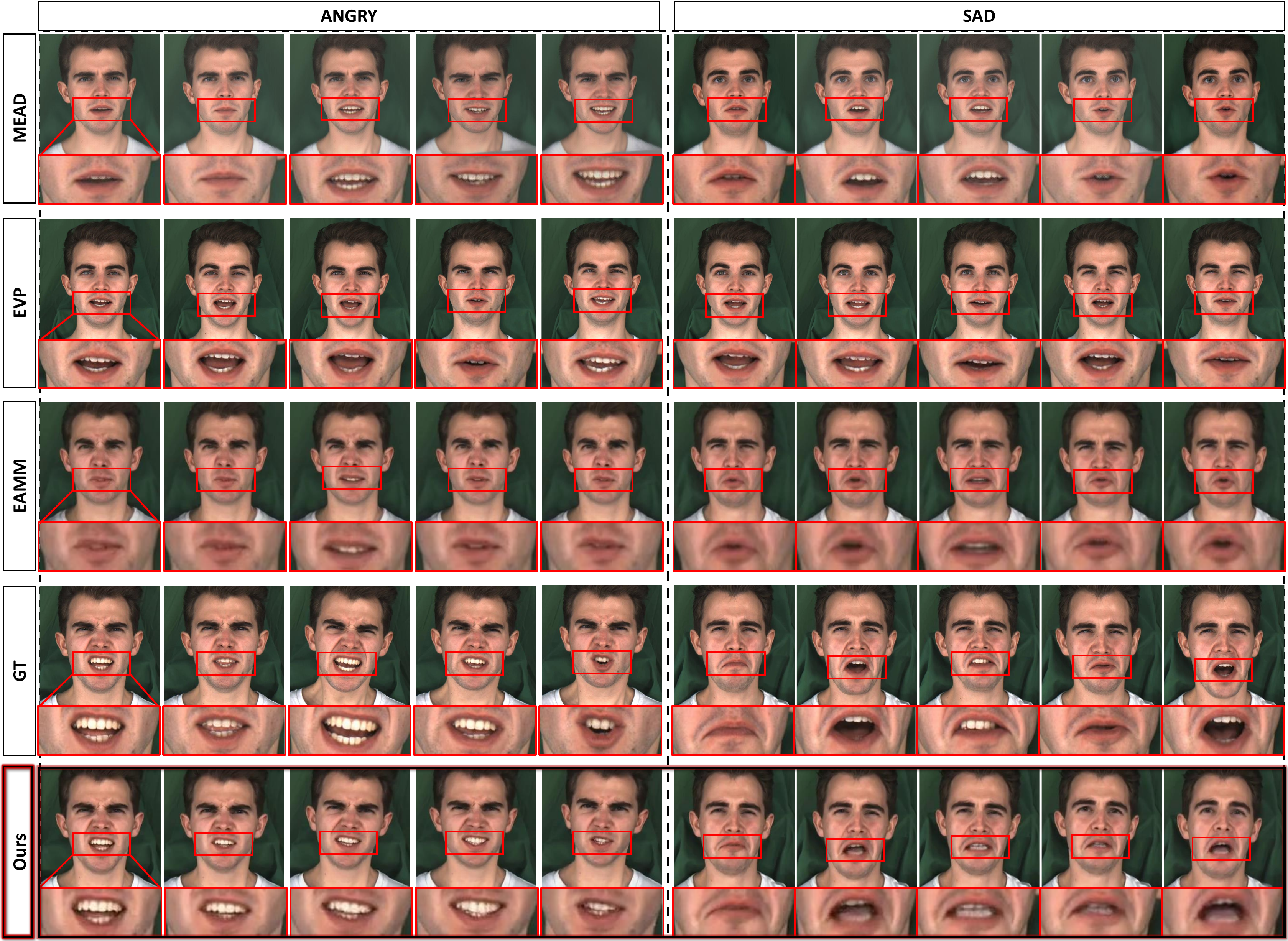}
    \caption{Qualitative comparison of our model and other state-of-the-art models for emotional talking face generation.}
    \vspace{-3mm}
    \label{fig:qualitative_compare}
\end{figure*}

\begin{table}[t]
  \caption{Performance of ablation study in DTW, including the absence of Gaze Sequentializer, denoted as Ours w/o Gaze, and the absence of Pose Sequentializer, denoted as Ours w/o Pose.}
  \label{tab:Ablation}
  \scriptsize%
	\centering%
 \resizebox{0.9\linewidth}{!}{
  \begin{tabu}{l|ccccc}
  \toprule
   Method & Pitch $\downarrow$ &  Yaw $\downarrow$  & Roll $\downarrow$ & Left eye $\downarrow$ & Right eye $\downarrow$\\
  \midrule
   MEAD\cite{MEAD} & 0.62 & 0.32 & 0.53 & 0.85 & 0.91\\
   EVP\cite{EVP}  & 0.74 & 0.77 & 0.62 & 0.88 & 0.95\\
   EAMM\cite{EAMM} & 0.76 & 0.74 & 0.31 & 0.81 & 0.88\\
   \midrule
   Ours w/o Gaze & 0.58 & 0.38 & 0.33 & 0.83 & 0.89\\
   Ours w/o Pose & 0.70 & 0.65 & 0.42 & 0.68 & 0.71\\
   Ours w/o C    & 0.55 & 0.42 & 0.27 & 0.64 & 0.70\\
   Ours     
   & \textbf{0.53} & \textbf{0.31} & \textbf{0.18} & \textbf{0.62}  & \textbf{0.68} \\
  \bottomrule
  \end{tabu}%
  }
\end{table}

\subsection{Evaluation of Facial Cues.}
\noindent\textbf{Evaluation of Normalized Landmarks.}
The accuracy of the generated facial landmarks critically influences the alignment of lip movements with the driving audio and the congruence between expressions and the corresponding emotion labels.
The normalization of facial landmarks removes the influence of different head poses on the movement direction of facial landmarks, allowing our method to focus on predicting the vertical motions of the landmarks in the mouth and eye regions, which is crucial for prediction accuracy.
We provide a qualitative comparison between the proposed Landmark Sequentializer and state-of-the-art methods for emotional talking face generation in Fig. \ref{fig:lms_compare}.
We found that only our method can effectively predict eye blinks and generate lips with better alignment (see the red boxes), which is consistent with the landmarks errors presented in Tab. \ref{tab:quantitative}

\begin{table}[t]
  \caption{Quantitative results of different talking face generation models on MEAD dataset.}
  \label{tab:quantitative}
  \scriptsize%
	\centering%
 \resizebox{\linewidth}{!}{
  \begin{tabu}{l|cccccc}
  \toprule
   Method & MLD $\downarrow$ &  FLD $\downarrow$  & SynNet $\uparrow$ & SSIM $\uparrow$ & PSNR $\uparrow$ & FID $\downarrow$ \\
  \midrule
   ATVG\cite{chen2019hierarchical} 
   & 3.14 & 3.87 & 2.24 & 0.57 & 28.58 & 67.6 \\
   SDA\cite{vougioukas2018end}
   & 3.99 & 4.5 & 1.88 & 0.44 & 28.54 & - \\
   Wave2Lip\cite{prajwal2020lip}
   & 3.43 & 3.80 & 2.24 & 0.57 & 29.03 & - \\
   MakeItTalk\cite{zhou2020makelttalk}
   & 3.80 & 3.92 & 2.20 & 0.56 & 28.92 & - \\
   PC-AVS\cite{zhou2021pose}
   & 2.97 & 2.74 & 2.10 & 0.60 & 29.02 & - \\
   Song\cite{song2022everybody}       
   & 2.54 & 3.49 & - & 0.64 & 29.11 & 36.33 \\
     \midrule
   MEAD\cite{MEAD}
   & 2.52 & 3.16 & - & 0.68 & 28.61 & 22.52 \\
   EVP\cite{EVP} 
   & 2.45 & 3.01 & - & 0.71 & 29.53 & \textbf{7.99} \\
   EAMM\cite{EAMM}
   & 2.41 & 2.55 & 2.26 & 0.66 & 29.29 & - \\
   Xu\cite{xu2023high}
   & 2.31 & - & 3.57 & \textbf{0.75} & 30.10 & 15.89 \\
   EMMN\cite{tan2023emmn}
   & 2.78 & 2.87 & 3.57 & 0.66 & 29.38 & - \\
     \midrule
   Ours w/o C
   & \underline{2.21} & \underline{2.11} & \underline{4.53} & 0.69 & \underline{30.14} & 9.12 \\
   Ours     
   & \textbf{2.08} & \textbf{1.99} & \textbf{4.72} & \underline{0.74} & \textbf{30.98} & \underline{8.62} \\
  \bottomrule
  \end{tabu}%
  }
\end{table}

\noindent\textbf{Evaluation of Head Pose.}
The generation and evaluation of head pose sequences are still challenges in this task of audio-driven talking face generation.
Due to the complex relationship between driving audio and the resultant head pose sequences, which are not mapped one-to-one, quantitatively assessing head pose generation is inherently difficult. 
Therefore, we randomly selected generated videos across various emotional categories. Then, we plotted the pitch, yaw, and roll of the generated head pose sequences and corresponding ground truth (GT) as line charts, as shown in Fig. \ref{fig:he_res}.
Across different emotion categories, the generated head pose sequences exhibit similar trends to the ground truth (GT), demonstrating that the head poses generated by our proposed method can translate into emotionally aligned head movements. 
Additionally, the quantitative results presented in Tab. \ref{tab:Ablation} corroborate the effectiveness of the Pose Sequentializer, consistent with trends shown in Fig. \ref{fig:he_res}.

\noindent\textbf{Evaluation of Gaze.}
The direction of gaze is a critical facial cue in emotional talking face generation, significantly influencing the vividness of the generated video.
Based on the gaze discretization strategy mentioned in Sec. \ref{sec:data_preprocess}, we verify the effectiveness of the proposed Gaze Sequentializer by analyzing the spatial distribution of the pupils in the eye region.
As shown in Fig. \ref{fig:gaze_res}, pie charts were used to represent the gaze distribution of different methods across various emotion categories.
We found that MEAD \cite{MEAD} and EAMM \cite{EAMM} exhibit identical gaze distributions across different emotion categories, with the majority of the pupils falling within the eye region "5". It implies that these two methods are incapable of generating available gaze sequences.
The EVP \cite{EVP} can only generate gaze distributions similar to the ground truth (GT) in the "happy" category, while it produces random gaze distributions in other emotion categories.
Our method is capable of generating gaze distributions similar to the GT across all categories of emotions.
In addition, we have further validated the effectiveness of the Gaze Sequentializer by calculating the DTW of pupil movement speed sequences as shown in Tab. \ref{tab:Ablation}.
These results demonstrate that the gaze sequences generated by our method effectively align with the corresponding emotional labels.
To the best of our knowledge, we are the first speech-driven talking face generation method capable of generating vivid gaze sequences related to emotion categories.

\subsection{Comparison with State-of-the-art Methods.}
\noindent\textbf{Quantitative Evaluation.}
To conduct a comprehensive evaluation, we performed quantitative comparisons between our method and other approaches in both emotional and non-emotional talking face generation.
Tab. \ref{tab:quantitative} reports the quantitative experimental results.
Apart from FID and SSIM, our method outperforms other approaches significantly in terms of MLD, FLD, and the confidence of SynNet, owing to the joint contributions from all three modules we proposed.
The notable performance of EVP \cite{EVP} on the FID metric primarily results from their substantial investment in independently training face generation models for each individual.
The improvements in MLD, FLD and SynNet shows that our approach can generate more accurate facial landmarks and achieve better lip alignment, consistent with the qualitative results shown in Fig. \ref{fig:lms_compare}.
The results of SSIM, PSNR, and FID further confirm that the proposed method can generate high-quality facial images.
We also conducted an ablation study to validate the efficacy of the collaborative emotion classifier.
The results, as presented in Tab. \ref{tab:quantitative} and \ref{tab:Ablation}, demonstrate that the absence of the emotion classifier, denoted as 'Ours w/o C', results in a significant decline in all quantitative metrics.
This decrease is primarily attributed to the intrinsic connection established by the emotion classifier among landmarks, gaze, and head pose, enhancing their consistency.

\noindent\textbf{Qualitative Evaluation.}
Fig. \ref{fig:qualitative_compare} presents the qualitative comparison results of our method with other state-of-the-art emotional talking face generation methods.
Due to the high alignment between facial cues and emotion categories, our method significantly  outperforms other approaches in terms of normalized landmarks, gaze, and head poses.
Specifically, our method can generate lip shapes that are more consistent with the input speech, attributable to the high precision of normalized landmark generation. 
Additionally, the natural variations in head pose and gaze in the generated face sequences further demonstrate the effectiveness of the proposed method. 
Relevant comparison videos are provided in the supplementary materials.
\begin{figure}[t]
    \centering
    \includegraphics[width=1.0\columnwidth]{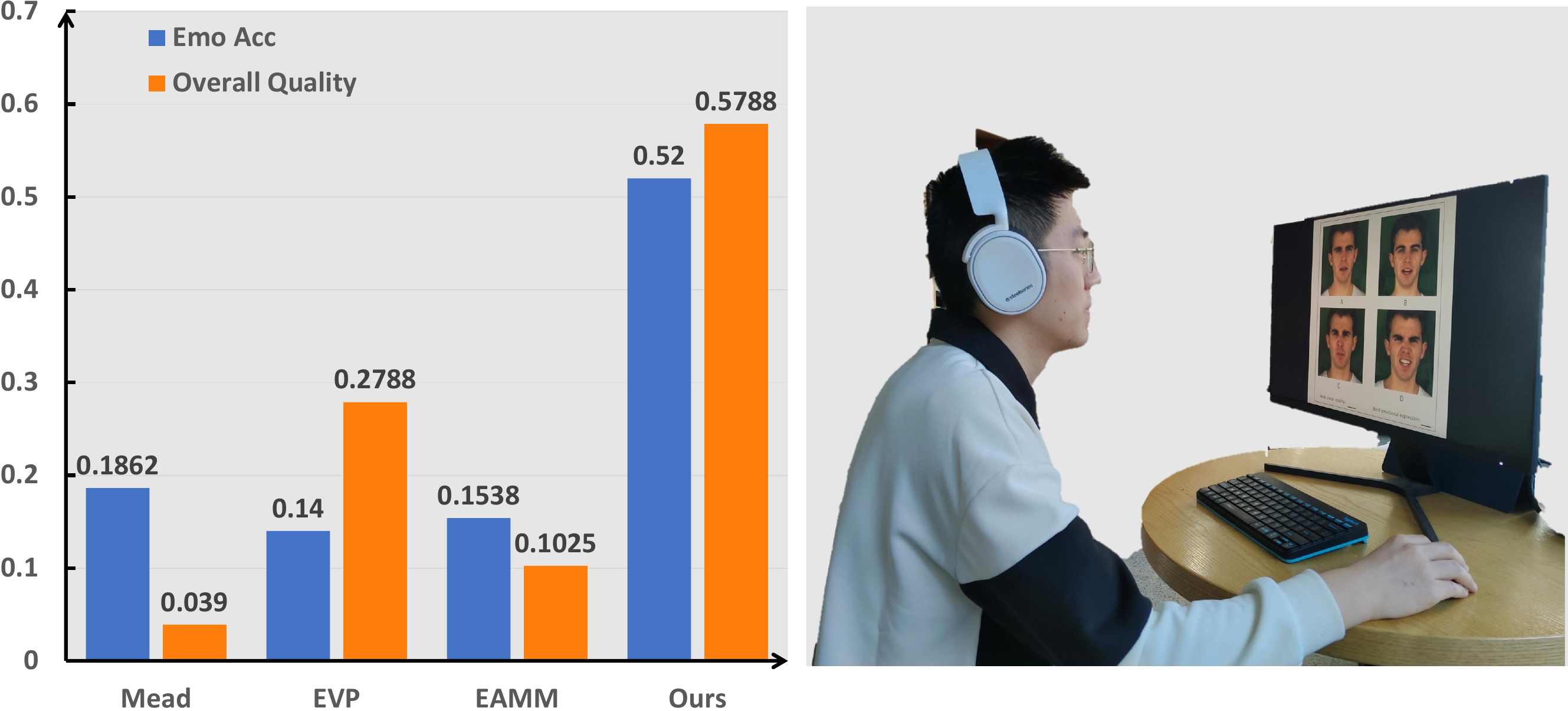}
    \caption{User study results of talking face video generation quality and emotion accuracy.}
    \vspace{-5mm}
    \label{fig:user_study}
\end{figure}

\noindent\textbf{User Study.}
We conducted a user study to compare our method with three SOTA models: MEAD, EVP, and EAMM.
Specifically, we randomly selected five videos from each emotion category in the MEAD test set and evaluated both the overall quality and the accuracy of emotion expression in the generated videos. Human subjects were asked to vote for the video with the highest overall quality and the most accurate emotional expression.
The rank-1 ratio for each method is presented in Fig. \ref{fig:user_study}.
Our method achieves 52\% emotion accuracy and 57.88\%  overall quality from 20 collected human subjects, significantly outperforming other methods.

\section{Conclusion}
We proposed a two-step framework to generate vivid talking faces with emotionally aligned facial cues.
In the first step, we aligned facial cues, including normalized facial landmarks, gaze, and head pose, with corresponding emotion labels in a self-supervised learning manner.
In the second step, we adopted latent key points as intermediate variables and utilized a pre-trained generative model to map various facial cues into high-quality facial images.
Extensive experiments on the MEAD dataset demonstrate that our model advances the state-of-the-art performance significantly.

\bibliographystyle{ACM-Reference-Format}
\bibliography{sample-base}


\begin{thebibliography}{38}


\ifx \showCODEN    \undefined \def \showCODEN     #1{\unskip}     \fi
\ifx \showDOI      \undefined \def \showDOI       #1{#1}\fi
\ifx \showISBNx    \undefined \def \showISBNx     #1{\unskip}     \fi
\ifx \showISBNxiii \undefined \def \showISBNxiii  #1{\unskip}     \fi
\ifx \showISSN     \undefined \def \showISSN      #1{\unskip}     \fi
\ifx \showLCCN     \undefined \def \showLCCN      #1{\unskip}     \fi
\ifx \shownote     \undefined \def \shownote      #1{#1}          \fi
\ifx \showarticletitle \undefined \def \showarticletitle #1{#1}   \fi
\ifx \showURL      \undefined \def \showURL       {\relax}        \fi
\providecommand\bibfield[2]{#2}
\providecommand\bibinfo[2]{#2}
\providecommand\natexlab[1]{#1}
\providecommand\showeprint[2][]{arXiv:#2}

\bibitem[Burkov et~al\mbox{.}(2020)]%
        {burkov2020neural}
\bibfield{author}{\bibinfo{person}{Egor Burkov}, \bibinfo{person}{Igor Pasechnik}, \bibinfo{person}{Artur Grigorev}, {and} \bibinfo{person}{Victor Lempitsky}.} \bibinfo{year}{2020}\natexlab{}.
\newblock \showarticletitle{Neural head reenactment with latent pose descriptors}. In \bibinfo{booktitle}{\emph{CVPR}}. \bibinfo{pages}{13786--13795}.
\newblock


\bibitem[Chen et~al\mbox{.}(2020)]%
        {chen2020talking}
\bibfield{author}{\bibinfo{person}{Lele Chen}, \bibinfo{person}{Guofeng Cui}, \bibinfo{person}{Celong Liu}, \bibinfo{person}{Zhong Li}, \bibinfo{person}{Ziyi Kou}, \bibinfo{person}{Yi Xu}, {and} \bibinfo{person}{Chenliang Xu}.} \bibinfo{year}{2020}\natexlab{}.
\newblock \showarticletitle{Talking-head generation with rhythmic head motion}. In \bibinfo{booktitle}{\emph{ECCV}}. Springer, \bibinfo{pages}{35--51}.
\newblock


\bibitem[Chen et~al\mbox{.}(2018)]%
        {chen2018lip}
\bibfield{author}{\bibinfo{person}{Lele Chen}, \bibinfo{person}{Zhiheng Li}, \bibinfo{person}{Ross~K Maddox}, \bibinfo{person}{Zhiyao Duan}, {and} \bibinfo{person}{Chenliang Xu}.} \bibinfo{year}{2018}\natexlab{}.
\newblock \showarticletitle{Lip movements generation at a glance}. In \bibinfo{booktitle}{\emph{Proceedings of the European conference on computer vision (ECCV)}}. \bibinfo{pages}{520--535}.
\newblock


\bibitem[Chen et~al\mbox{.}(2019)]%
        {chen2019hierarchical}
\bibfield{author}{\bibinfo{person}{Lele Chen}, \bibinfo{person}{Ross~K Maddox}, \bibinfo{person}{Zhiyao Duan}, {and} \bibinfo{person}{Chenliang Xu}.} \bibinfo{year}{2019}\natexlab{}.
\newblock \showarticletitle{Hierarchical cross-modal talking face generation with dynamic pixel-wise loss}. In \bibinfo{booktitle}{\emph{CVPR}}. \bibinfo{pages}{7832--7841}.
\newblock


\bibitem[Chung and Zisserman(2017)]%
        {chung2017out}
\bibfield{author}{\bibinfo{person}{Joon~Son Chung} {and} \bibinfo{person}{Andrew Zisserman}.} \bibinfo{year}{2017}\natexlab{}.
\newblock \showarticletitle{Out of time: automated lip sync in the wild}. In \bibinfo{booktitle}{\emph{Workshop on Multi-view Lip-reading, ACCV.}} Springer, \bibinfo{pages}{251--263}.
\newblock


\bibitem[Drobyshev et~al\mbox{.}(2022)]%
        {drobyshev2022megaportraits}
\bibfield{author}{\bibinfo{person}{Nikita Drobyshev}, \bibinfo{person}{Jenya Chelishev}, \bibinfo{person}{Taras Khakhulin}, \bibinfo{person}{Aleksei Ivakhnenko}, \bibinfo{person}{Victor Lempitsky}, {and} \bibinfo{person}{Egor Zakharov}.} \bibinfo{year}{2022}\natexlab{}.
\newblock \showarticletitle{Megaportraits: One-shot megapixel neural head avatars}. In \bibinfo{booktitle}{\emph{ACM MM}}. \bibinfo{pages}{2663--2671}.
\newblock


\bibitem[Du et~al\mbox{.}(2023)]%
        {du2023dae}
\bibfield{author}{\bibinfo{person}{Chenpeng Du}, \bibinfo{person}{Qi Chen}, \bibinfo{person}{Tianyu He}, \bibinfo{person}{Xu Tan}, \bibinfo{person}{Xie Chen}, \bibinfo{person}{Kai Yu}, \bibinfo{person}{Sheng Zhao}, {and} \bibinfo{person}{Jiang Bian}.} \bibinfo{year}{2023}\natexlab{}.
\newblock \showarticletitle{Dae-talker: High fidelity speech-driven talking face generation with diffusion autoencoder}. In \bibinfo{booktitle}{\emph{Proceedings of the 31st ACM International Conference on Multimedia}}. \bibinfo{pages}{4281--4289}.
\newblock


\bibitem[Fan et~al\mbox{.}(2022)]%
        {fan2022faceformer}
\bibfield{author}{\bibinfo{person}{Yingruo Fan}, \bibinfo{person}{Zhaojiang Lin}, \bibinfo{person}{Jun Saito}, \bibinfo{person}{Wenping Wang}, {and} \bibinfo{person}{Taku Komura}.} \bibinfo{year}{2022}\natexlab{}.
\newblock \showarticletitle{Faceformer: Speech-driven 3d facial animation with transformers}. In \bibinfo{booktitle}{\emph{CVPR}}. \bibinfo{pages}{18770--18780}.
\newblock


\bibitem[Gu et~al\mbox{.}(2020)]%
        {gu2020flnet}
\bibfield{author}{\bibinfo{person}{Kuangxiao Gu}, \bibinfo{person}{Yuqian Zhou}, {and} \bibinfo{person}{Thomas Huang}.} \bibinfo{year}{2020}\natexlab{}.
\newblock \showarticletitle{Flnet: Landmark driven fetching and learning network for faithful talking facial animation synthesis}. In \bibinfo{booktitle}{\emph{AAAI}}, Vol.~\bibinfo{volume}{34}. \bibinfo{pages}{10861--10868}.
\newblock


\bibitem[Guo et~al\mbox{.}(2020)]%
        {guo2020towards}
\bibfield{author}{\bibinfo{person}{Jianzhu Guo}, \bibinfo{person}{Xiangyu Zhu}, \bibinfo{person}{Yang Yang}, \bibinfo{person}{Fan Yang}, \bibinfo{person}{Zhen Lei}, {and} \bibinfo{person}{Stan~Z Li}.} \bibinfo{year}{2020}\natexlab{}.
\newblock \showarticletitle{Towards fast, accurate and stable 3d dense face alignment}. In \bibinfo{booktitle}{\emph{ECCV}}. Springer, \bibinfo{pages}{152--168}.
\newblock


\bibitem[Gururani et~al\mbox{.}(2023)]%
        {space}
\bibfield{author}{\bibinfo{person}{Siddharth Gururani}, \bibinfo{person}{Arun Mallya}, \bibinfo{person}{Ting-Chun Wang}, \bibinfo{person}{Rafael Valle}, {and} \bibinfo{person}{Ming-Yu Liu}.} \bibinfo{year}{2023}\natexlab{}.
\newblock \showarticletitle{Space: Speech-driven portrait animation with controllable expression}. In \bibinfo{booktitle}{\emph{Proceedings of the IEEE/CVF International Conference on Computer Vision}}. \bibinfo{pages}{20914--20923}.
\newblock


\bibitem[Heusel et~al\mbox{.}(2017)]%
        {heusel2017gans}
\bibfield{author}{\bibinfo{person}{Martin Heusel}, \bibinfo{person}{Hubert Ramsauer}, \bibinfo{person}{Thomas Unterthiner}, \bibinfo{person}{Bernhard Nessler}, {and} \bibinfo{person}{Sepp Hochreiter}.} \bibinfo{year}{2017}\natexlab{}.
\newblock \showarticletitle{Gans trained by a two time-scale update rule converge to a local nash equilibrium}.
\newblock \bibinfo{journal}{\emph{NIPS}}  \bibinfo{volume}{30} (\bibinfo{year}{2017}).
\newblock


\bibitem[Hong et~al\mbox{.}(2022)]%
        {hong2022depth}
\bibfield{author}{\bibinfo{person}{Fa-Ting Hong}, \bibinfo{person}{Longhao Zhang}, \bibinfo{person}{Li Shen}, {and} \bibinfo{person}{Dan Xu}.} \bibinfo{year}{2022}\natexlab{}.
\newblock \showarticletitle{Depth-aware generative adversarial network for talking head video generation}. In \bibinfo{booktitle}{\emph{CVPR}}. \bibinfo{pages}{3397--3406}.
\newblock


\bibitem[Ji et~al\mbox{.}(2022)]%
        {EAMM}
\bibfield{author}{\bibinfo{person}{Xinya Ji}, \bibinfo{person}{Hang Zhou}, \bibinfo{person}{Kaisiyuan Wang}, \bibinfo{person}{Qianyi Wu}, \bibinfo{person}{Wayne Wu}, \bibinfo{person}{Feng Xu}, {and} \bibinfo{person}{Xun Cao}.} \bibinfo{year}{2022}\natexlab{}.
\newblock \showarticletitle{Eamm: One-shot emotional talking face via audio-based emotion-aware motion model}. In \bibinfo{booktitle}{\emph{ACM SIGGRAPH}}. \bibinfo{pages}{1--10}.
\newblock


\bibitem[Ji et~al\mbox{.}(2021)]%
        {EVP}
\bibfield{author}{\bibinfo{person}{Xinya Ji}, \bibinfo{person}{Hang Zhou}, \bibinfo{person}{Kaisiyuan Wang}, \bibinfo{person}{Wayne Wu}, \bibinfo{person}{Chen~Change Loy}, \bibinfo{person}{Xun Cao}, {and} \bibinfo{person}{Feng Xu}.} \bibinfo{year}{2021}\natexlab{}.
\newblock \showarticletitle{Audio-driven emotional video portraits}. In \bibinfo{booktitle}{\emph{CVPR}}. \bibinfo{pages}{14080--14089}.
\newblock


\bibitem[Karras et~al\mbox{.}(2017)]%
        {karras2017audio}
\bibfield{author}{\bibinfo{person}{Tero Karras}, \bibinfo{person}{Timo Aila}, \bibinfo{person}{Samuli Laine}, \bibinfo{person}{Antti Herva}, {and} \bibinfo{person}{Jaakko Lehtinen}.} \bibinfo{year}{2017}\natexlab{}.
\newblock \showarticletitle{Audio-driven facial animation by joint end-to-end learning of pose and emotion}.
\newblock \bibinfo{journal}{\emph{TOG}} \bibinfo{volume}{36}, \bibinfo{number}{4} (\bibinfo{year}{2017}), \bibinfo{pages}{1--12}.
\newblock


\bibitem[Liu et~al\mbox{.}(2021)]%
        {liu2021li}
\bibfield{author}{\bibinfo{person}{Jin Liu}, \bibinfo{person}{Peng Chen}, \bibinfo{person}{Tao Liang}, \bibinfo{person}{Zhaoxing Li}, \bibinfo{person}{Cai Yu}, \bibinfo{person}{Shuqiao Zou}, \bibinfo{person}{Jiao Dai}, {and} \bibinfo{person}{Jizhong Han}.} \bibinfo{year}{2021}\natexlab{}.
\newblock \showarticletitle{Li-net: Large-pose identity-preserving face reenactment network}. In \bibinfo{booktitle}{\emph{ICME}}. IEEE, \bibinfo{pages}{1--6}.
\newblock


\bibitem[Liu et~al\mbox{.}(2022)]%
        {liu2022semantic}
\bibfield{author}{\bibinfo{person}{Xian Liu}, \bibinfo{person}{Yinghao Xu}, \bibinfo{person}{Qianyi Wu}, \bibinfo{person}{Hang Zhou}, \bibinfo{person}{Wayne Wu}, {and} \bibinfo{person}{Bolei Zhou}.} \bibinfo{year}{2022}\natexlab{}.
\newblock \showarticletitle{Semantic-aware implicit neural audio-driven video portrait generation}. In \bibinfo{booktitle}{\emph{European Conference on Computer Vision}}. Springer, \bibinfo{pages}{106--125}.
\newblock


\bibitem[Lugaresi et~al\mbox{.}(2019)]%
        {lugaresi2019mediapipe}
\bibfield{author}{\bibinfo{person}{Camillo Lugaresi}, \bibinfo{person}{Jiuqiang Tang}, \bibinfo{person}{Hadon Nash}, \bibinfo{person}{Chris McClanahan}, \bibinfo{person}{Esha Uboweja}, \bibinfo{person}{Michael Hays}, \bibinfo{person}{Fan Zhang}, \bibinfo{person}{Chuo-Ling Chang}, \bibinfo{person}{Ming~Guang Yong}, \bibinfo{person}{Juhyun Lee}, {et~al\mbox{.}}} \bibinfo{year}{2019}\natexlab{}.
\newblock \showarticletitle{Mediapipe: A framework for building perception pipelines}.
\newblock \bibinfo{journal}{\emph{arXiv preprint arXiv:1906.08172}} (\bibinfo{year}{2019}).
\newblock


\bibitem[Prajwal et~al\mbox{.}(2020)]%
        {prajwal2020lip}
\bibfield{author}{\bibinfo{person}{KR Prajwal}, \bibinfo{person}{Rudrabha Mukhopadhyay}, \bibinfo{person}{Vinay~P Namboodiri}, {and} \bibinfo{person}{CV Jawahar}.} \bibinfo{year}{2020}\natexlab{}.
\newblock \showarticletitle{A lip sync expert is all you need for speech to lip generation in the wild}. In \bibinfo{booktitle}{\emph{ACM MM}}. \bibinfo{pages}{484--492}.
\newblock


\bibitem[Shen et~al\mbox{.}(2023)]%
        {shen2023difftalk}
\bibfield{author}{\bibinfo{person}{Shuai Shen}, \bibinfo{person}{Wenliang Zhao}, \bibinfo{person}{Zibin Meng}, \bibinfo{person}{Wanhua Li}, \bibinfo{person}{Zheng Zhu}, \bibinfo{person}{Jie Zhou}, {and} \bibinfo{person}{Jiwen Lu}.} \bibinfo{year}{2023}\natexlab{}.
\newblock \showarticletitle{DiffTalk: Crafting Diffusion Models for Generalized Audio-Driven Portraits Animation}. In \bibinfo{booktitle}{\emph{Proceedings of the IEEE/CVF Conference on Computer Vision and Pattern Recognition}}. \bibinfo{pages}{1982--1991}.
\newblock


\bibitem[Siarohin et~al\mbox{.}(2021)]%
        {siarohin2021motion}
\bibfield{author}{\bibinfo{person}{Aliaksandr Siarohin}, \bibinfo{person}{Oliver~J Woodford}, \bibinfo{person}{Jian Ren}, \bibinfo{person}{Menglei Chai}, {and} \bibinfo{person}{Sergey Tulyakov}.} \bibinfo{year}{2021}\natexlab{}.
\newblock \showarticletitle{Motion representations for articulated animation}. In \bibinfo{booktitle}{\emph{CVPR}}. \bibinfo{pages}{13653--13662}.
\newblock


\bibitem[Song et~al\mbox{.}(2022)]%
        {song2022everybody}
\bibfield{author}{\bibinfo{person}{Linsen Song}, \bibinfo{person}{Wayne Wu}, \bibinfo{person}{Chen Qian}, \bibinfo{person}{Ran He}, {and} \bibinfo{person}{Chen~Change Loy}.} \bibinfo{year}{2022}\natexlab{}.
\newblock \showarticletitle{Everybody’s talkin’: Let me talk as you want}.
\newblock \bibinfo{journal}{\emph{IEEE Transactions on Information Forensics and Security}}  \bibinfo{volume}{17} (\bibinfo{year}{2022}), \bibinfo{pages}{585--598}.
\newblock


\bibitem[Tan et~al\mbox{.}(2023)]%
        {tan2023emmn}
\bibfield{author}{\bibinfo{person}{Shuai Tan}, \bibinfo{person}{Bin Ji}, {and} \bibinfo{person}{Ye Pan}.} \bibinfo{year}{2023}\natexlab{}.
\newblock \showarticletitle{Emmn: Emotional motion memory network for audio-driven emotional talking face generation}. In \bibinfo{booktitle}{\emph{Proceedings of the IEEE/CVF International Conference on Computer Vision}}. \bibinfo{pages}{22146--22156}.
\newblock


\bibitem[Taylor et~al\mbox{.}(2017)]%
        {taylor2017deep}
\bibfield{author}{\bibinfo{person}{Sarah Taylor}, \bibinfo{person}{Taehwan Kim}, \bibinfo{person}{Yisong Yue}, \bibinfo{person}{Moshe Mahler}, \bibinfo{person}{James Krahe}, \bibinfo{person}{Anastasio~Garcia Rodriguez}, \bibinfo{person}{Jessica Hodgins}, {and} \bibinfo{person}{Iain Matthews}.} \bibinfo{year}{2017}\natexlab{}.
\newblock \showarticletitle{A deep learning approach for generalized speech animation}.
\newblock \bibinfo{journal}{\emph{TOG}} \bibinfo{volume}{36}, \bibinfo{number}{4} (\bibinfo{year}{2017}), \bibinfo{pages}{1--11}.
\newblock


\bibitem[Vougioukas et~al\mbox{.}(2019)]%
        {vougioukas2018end}
\bibfield{author}{\bibinfo{person}{Konstantinos Vougioukas}, \bibinfo{person}{Stavros Petridis}, {and} \bibinfo{person}{Maja Pantic}.} \bibinfo{year}{2019}\natexlab{}.
\newblock \showarticletitle{End-to-end speech-driven facial animation with temporal GANs}. In \bibinfo{booktitle}{\emph{CVPR Workshop}}.
\newblock


\bibitem[Wang et~al\mbox{.}(2023a)]%
        {PDFC}
\bibfield{author}{\bibinfo{person}{Duomin Wang}, \bibinfo{person}{Yu Deng}, \bibinfo{person}{Zixin Yin}, \bibinfo{person}{Heung-Yeung Shum}, {and} \bibinfo{person}{Baoyuan Wang}.} \bibinfo{year}{2023}\natexlab{a}.
\newblock \showarticletitle{Progressive Disentangled Representation Learning for Fine-Grained Controllable Talking Head Synthesis}. In \bibinfo{booktitle}{\emph{CVPR}}. \bibinfo{pages}{17979--17989}.
\newblock


\bibitem[Wang et~al\mbox{.}(2023b)]%
        {wang2023lipformer}
\bibfield{author}{\bibinfo{person}{Jiayu Wang}, \bibinfo{person}{Kang Zhao}, \bibinfo{person}{Shiwei Zhang}, \bibinfo{person}{Yingya Zhang}, \bibinfo{person}{Yujun Shen}, \bibinfo{person}{Deli Zhao}, {and} \bibinfo{person}{Jingren Zhou}.} \bibinfo{year}{2023}\natexlab{b}.
\newblock \showarticletitle{LipFormer: High-Fidelity and Generalizable Talking Face Generation With a Pre-Learned Facial Codebook}. In \bibinfo{booktitle}{\emph{Proceedings of the IEEE/CVF Conference on Computer Vision and Pattern Recognition}}. \bibinfo{pages}{13844--13853}.
\newblock


\bibitem[Wang et~al\mbox{.}(2020)]%
        {MEAD}
\bibfield{author}{\bibinfo{person}{Kaisiyuan Wang}, \bibinfo{person}{Qianyi Wu}, \bibinfo{person}{Linsen Song}, \bibinfo{person}{Zhuoqian Yang}, \bibinfo{person}{Wayne Wu}, \bibinfo{person}{Chen Qian}, \bibinfo{person}{Ran He}, \bibinfo{person}{Yu Qiao}, {and} \bibinfo{person}{Chen~Change Loy}.} \bibinfo{year}{2020}\natexlab{}.
\newblock \showarticletitle{Mead: A large-scale audio-visual dataset for emotional talking-face generation}. In \bibinfo{booktitle}{\emph{ECCV}}. Springer, \bibinfo{pages}{700--717}.
\newblock


\bibitem[Wang et~al\mbox{.}(2021a)]%
        {wang2021audio2head}
\bibfield{author}{\bibinfo{person}{Suzhen Wang}, \bibinfo{person}{Lincheng Li}, \bibinfo{person}{Yu Ding}, \bibinfo{person}{Changjie Fan}, {and} \bibinfo{person}{Xin Yu}.} \bibinfo{year}{2021}\natexlab{a}.
\newblock \showarticletitle{Audio2head: Audio-driven one-shot talking-head generation with natural head motion}.
\newblock \bibinfo{journal}{\emph{IJCAI}} (\bibinfo{year}{2021}).
\newblock


\bibitem[Wang et~al\mbox{.}(2021b)]%
        {facevid2vid}
\bibfield{author}{\bibinfo{person}{Ting-Chun Wang}, \bibinfo{person}{Arun Mallya}, {and} \bibinfo{person}{Ming-Yu Liu}.} \bibinfo{year}{2021}\natexlab{b}.
\newblock \showarticletitle{One-shot free-view neural talking-head synthesis for video conferencing}. In \bibinfo{booktitle}{\emph{CVPR}}. \bibinfo{pages}{10039--10049}.
\newblock


\bibitem[Wang et~al\mbox{.}(2022)]%
        {wang2022latent}
\bibfield{author}{\bibinfo{person}{Yaohui Wang}, \bibinfo{person}{Di Yang}, \bibinfo{person}{Francois Bremond}, {and} \bibinfo{person}{Antitza Dantcheva}.} \bibinfo{year}{2022}\natexlab{}.
\newblock \showarticletitle{Latent image animator: Learning to animate images via latent space navigation}.
\newblock \bibinfo{journal}{\emph{ICLR}}.
\newblock


\bibitem[Wu et~al\mbox{.}(2023)]%
        {wu2023speech}
\bibfield{author}{\bibinfo{person}{Haozhe Wu}, \bibinfo{person}{Songtao Zhou}, \bibinfo{person}{Jia Jia}, \bibinfo{person}{Junliang Xing}, \bibinfo{person}{Qi Wen}, {and} \bibinfo{person}{Xiang Wen}.} \bibinfo{year}{2023}\natexlab{}.
\newblock \showarticletitle{Speech-Driven 3D Face Animation with Composite and Regional Facial Movements}. In \bibinfo{booktitle}{\emph{Proceedings of the 31st ACM International Conference on Multimedia}}. \bibinfo{pages}{6822--6830}.
\newblock


\bibitem[Xiang et~al\mbox{.}(2020)]%
        {xiang2020one}
\bibfield{author}{\bibinfo{person}{Sitao Xiang}, \bibinfo{person}{Yuming Gu}, \bibinfo{person}{Pengda Xiang}, \bibinfo{person}{Mingming He}, \bibinfo{person}{Koki Nagano}, \bibinfo{person}{Haiwei Chen}, {and} \bibinfo{person}{Hao Li}.} \bibinfo{year}{2020}\natexlab{}.
\newblock \showarticletitle{One-shot identity-preserving portrait reenactment}.
\newblock \bibinfo{journal}{\emph{arXiv preprint arXiv:2004.12452}} (\bibinfo{year}{2020}).
\newblock


\bibitem[Xu et~al\mbox{.}(2023)]%
        {xu2023high}
\bibfield{author}{\bibinfo{person}{Chao Xu}, \bibinfo{person}{Junwei Zhu}, \bibinfo{person}{Jiangning Zhang}, \bibinfo{person}{Yue Han}, \bibinfo{person}{Wenqing Chu}, \bibinfo{person}{Ying Tai}, \bibinfo{person}{Chengjie Wang}, \bibinfo{person}{Zhifeng Xie}, {and} \bibinfo{person}{Yong Liu}.} \bibinfo{year}{2023}\natexlab{}.
\newblock \showarticletitle{High-fidelity Generalized Emotional Talking Face Generation with Multi-modal Emotion Space Learning}. In \bibinfo{booktitle}{\emph{Proceedings of the IEEE/CVF Conference on Computer Vision and Pattern Recognition}}. \bibinfo{pages}{6609--6619}.
\newblock


\bibitem[Zhou et~al\mbox{.}(2019)]%
        {zhou2019talking}
\bibfield{author}{\bibinfo{person}{Hang Zhou}, \bibinfo{person}{Yu Liu}, \bibinfo{person}{Ziwei Liu}, \bibinfo{person}{Ping Luo}, {and} \bibinfo{person}{Xiaogang Wang}.} \bibinfo{year}{2019}\natexlab{}.
\newblock \showarticletitle{Talking face generation by adversarially disentangled audio-visual representation}. In \bibinfo{booktitle}{\emph{AAAI}}, Vol.~\bibinfo{volume}{33}. \bibinfo{pages}{9299--9306}.
\newblock


\bibitem[Zhou et~al\mbox{.}(2021)]%
        {zhou2021pose}
\bibfield{author}{\bibinfo{person}{Hang Zhou}, \bibinfo{person}{Yasheng Sun}, \bibinfo{person}{Wayne Wu}, \bibinfo{person}{Chen~Change Loy}, \bibinfo{person}{Xiaogang Wang}, {and} \bibinfo{person}{Ziwei Liu}.} \bibinfo{year}{2021}\natexlab{}.
\newblock \showarticletitle{Pose-controllable talking face generation by implicitly modularized audio-visual representation}. In \bibinfo{booktitle}{\emph{CVPR}}. \bibinfo{pages}{4176--4186}.
\newblock


\bibitem[Zhou et~al\mbox{.}(2020)]%
        {zhou2020makelttalk}
\bibfield{author}{\bibinfo{person}{Yang Zhou}, \bibinfo{person}{Xintong Han}, \bibinfo{person}{Eli Shechtman}, \bibinfo{person}{Jose Echevarria}, \bibinfo{person}{Evangelos Kalogerakis}, {and} \bibinfo{person}{Dingzeyu Li}.} \bibinfo{year}{2020}\natexlab{}.
\newblock \showarticletitle{Makelttalk: speaker-aware talking-head animation}.
\newblock \bibinfo{journal}{\emph{TOG}} \bibinfo{volume}{39}, \bibinfo{number}{6} (\bibinfo{year}{2020}), \bibinfo{pages}{1--15}.
\newblock


\end{thebibliography}

\end{document}